\documentclass[10pt,twocolumn,letterpaper]{article}

\usepackage{cvpr}
\usepackage{times}
\usepackage{epsfig}
\usepackage{graphicx}
\usepackage{subfigure}
\usepackage{amsmath}
\usepackage{amssymb}
\usepackage{algorithm}
\usepackage{array}
\usepackage{algorithmic}
\usepackage{bm}
\usepackage{tabularx}

\usepackage[breaklinks=true,bookmarks=false]{hyperref}

\cvprfinalcopy 

\def\cvprPaperID{****} 

\begin{document}

\title{Graph Pruning for Model Compression}

\author{Mingyang Zhang, Xinyi Yu, Jingtao Rong, Linlin Ou\\
College of Information Engineering, Zhejiang University of Technology\\
Hang Zhou, People’s Republic of China\\
{\tt\small linlinou@zjut.edu.cn}
}

\maketitle

\begin{abstract}
Previous AutoML pruning works utilized individual layer features to automatically prune filters. We analyze the correlation for two layers from the different blocks which have a short-cut structure. It shows that, in one block, the deeper layer has many redundant filters which can be represented by filters in the former layer. So, it is necessary to take information from other layers into consideration in pruning. In this paper, a novel pruning method, named GraphPruning, is proposed. Any series of the network is viewed as a graph. To automatically aggregate neighboring features for each node, a graph aggregator based on graph convolution networks(GCN) is designed. In the training stage, a PruningNet that is given aggregated node features generates reasonable weights for any size of the sub-network. Subsequently, the best configuration of the Pruned Network is searched by reinforcement learning. Different from previous work, we take the node features from a well-trained graph aggregator instead of the hand-craft features, as the states in reinforcement learning. Compared with other AutoML pruning works, our method has achieved the state-of-the-art under the same conditions on ImageNet-2012. 
\end{abstract}
\section{Introduction}
Deep convolution neural networks (Deep CNNs), such as ResNet, DenseNet, MobileNet\cite{b1,b2,b3,b4}, etc., bring about the outstanding performance of computer vision applications, including object classification and localization, pedestrian and car detection, traffic flow management. However, constrained by latency, energy and computation complexity, it is hard to apply above superior networks to mobile phones, augmented reality devices and autonomous cars. Therefore, it is necessary to apply deep CNN models that can overcome these constraints and keep high accuracy. \\
Recent developments in network pruning can be divided into two main categories: weight pruning and filter pruning.
\begin{figure}
\centering
\includegraphics[width=4in]{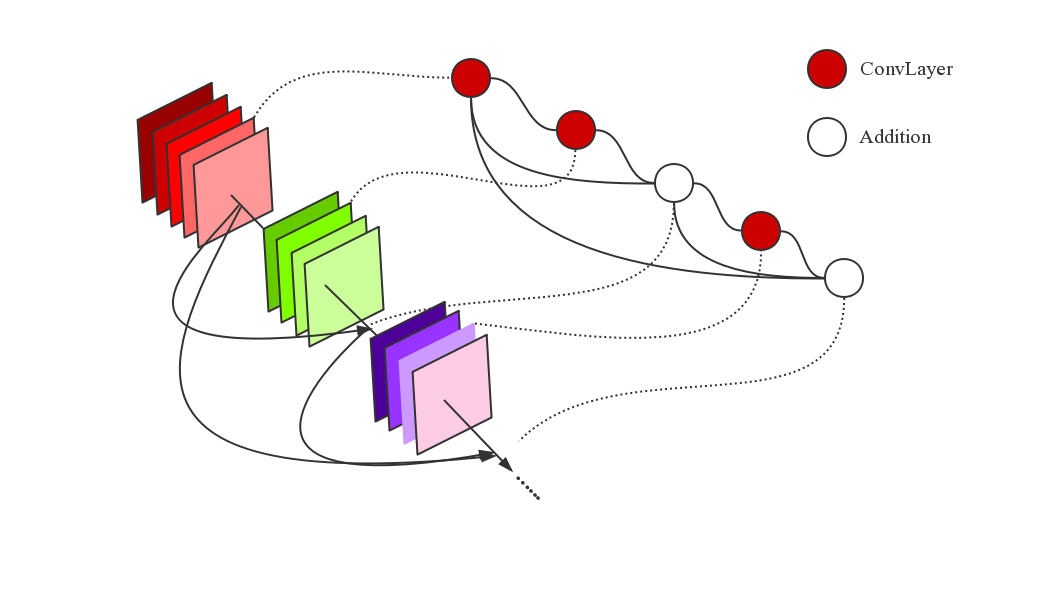}
\caption{We use a topology graph to describe any short-cut structure model. In the graph, each node represents a type of operation, such as normal convolution, depth-wise convolution, add and concatenation.}
\label{fig:topology}
\end{figure}
\begin{figure*}[htbp]
\subfigure[]{
\begin{minipage}[t]{0.3\linewidth}
\centering
\includegraphics[width=1.7in]{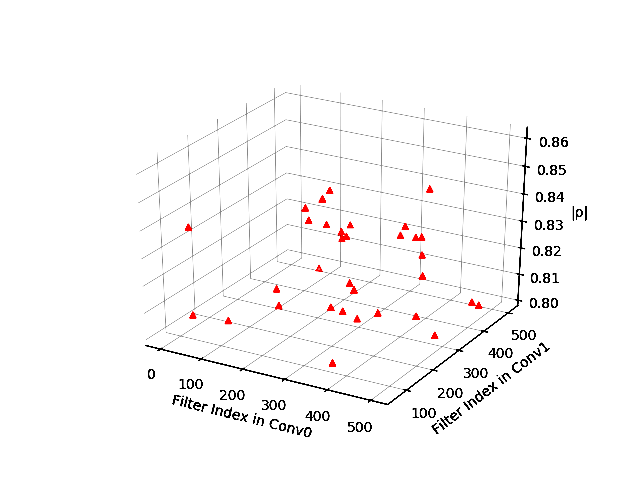}
\end{minipage}}
\subfigure[]{
\begin{minipage}[t]{0.3\linewidth}
\centering
\includegraphics[width=1.7in]{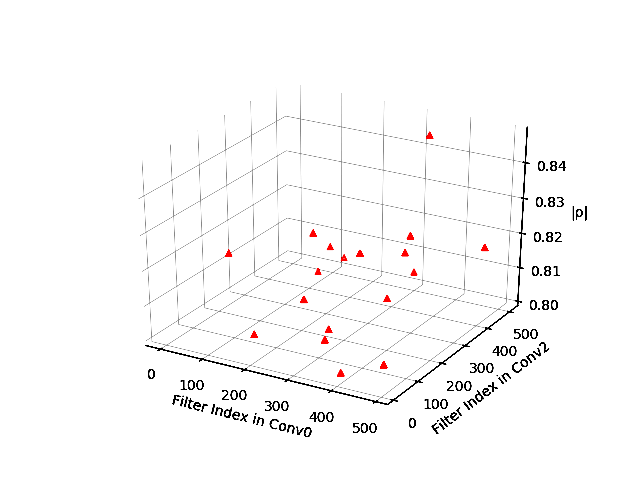}
\end{minipage}}
\subfigure[]{
\begin{minipage}[t]{0.3\linewidth}
\centering
\includegraphics[width=1.7in]{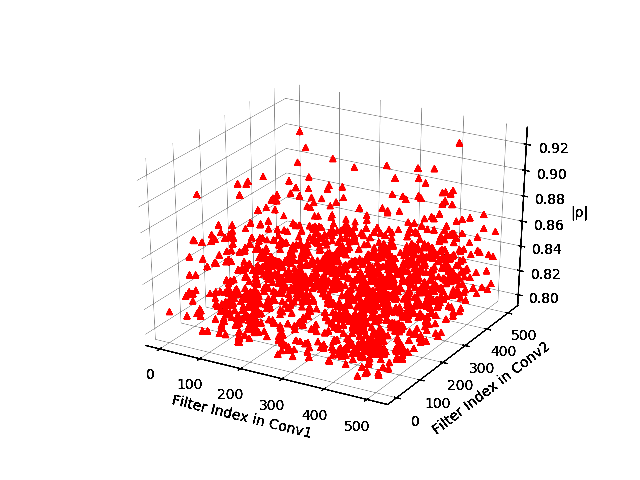}
\end{minipage}}
  \caption{$3\times 3$ convolution layers from the 4th stage in ResNet-50 are taken as an example. For the convenience of visualization, we compute the Pearson Correlation Matrix for convolution layer with $3\times 3$kernel size in each block and print pairs of filters which absolute Pearson Correlation value $|p|$ is larger than 0.8. (a) The filter correlation with 1st and 2nd convolution. (b) The filter correlation with 1st and 3rd convolution. (c) The filter correlation with 2nd and 3rd convolution.}
\label{fig:correlation}
\end{figure*}
Weight pruning directly removes the values of weights in a filer, which can result in unstructured sparsity. This irregular structure makes it difficult to utilize mainstream computational architecture \cite{b5}. Accordingly, filter pruning directly deletes the whole selected filter and rebuilds a narrow model with a regular structure. Therefore, filter pruning is more preferred to speed up the network and reduce the size of the model. Many recent works pay attention to filter pruning which directly discards the whole selected filters based on some human-defined rules \cite{b6,b52}. However, these rules cannot suit all models. For example, ResNet-50 has a residual connection while Vgg-19 \cite{b33} has not, so pruning these models based on the same rule is always sub-optimal. \\
To obtain a compressed model without human-defined rule and domain expertise, some automatically pruning methods are proposed \cite{b8,b9,b10}. Han \cite{b8} used reinforcement learning to automatically find pruning policy by the agent that takes features of each layer as input. The results of the reinforcement learning method surpass many human-craft pruning methods. Because the actor only obtains the features from an individual layer, it is still a lack of consideration for the overall network. Besides, it is not suitable for the large-scale dataset, such as ImageNet-2012 \cite{b15}. In evaluation, it needs to iteratively train to recover the accuracy. To prune networks on large-scale dataset, Liu \cite{b9} and Yu \cite{b10} proposed some one-shot architecture methods, which train a model that contains all sub-models. However, these methods just focus on one layer in pruning. We take the 4th stage from Resnet-50 as an example and use Pearson Correlation \cite{b12} to measure the similarity between each layer in different stages. As illustrated in \textbf{Fig. \ref{fig:correlation}}, the correlation between each layer is strong, especially for the layer with kernel size $3\times 3$, which means that some filters can be represented by filters from other layers. For a convolution layer of which input is produced by the former convolution layer, it may obtain a sub-optimal solution if only an individual layer information is considered.\\
To solve this problem, we propose a graph aggregator for model compression which converts the model to an undirected topology graph to find the correlation between each layer. In the training phase, the graph aggregator feed node features a PruningNet \cite{b9} which can automatically generate reasonable weights for the pruned layer. Thus, the graph aggregator and the PruningNet can be jointly trained on the dataset. Different from MetaPruning\cite{b9} that only provides one layer features for PruningNet, our method takes aggregated node features as input which gathers information of neighbor nodes. In the searching phase, we use a DDPG agent\cite{b13}  to search for the best configuration. Benefiting from the well-trained graph aggregator, the agent can take more advantage of the relationship between each layer. \\
\textbf{Contributions} There are three contributions:
\begin{itemize}
\item We demonstrate the importance of the filter relationship in pruning. The proposed GraphPruning method address this challenging problem.
\item We propose a graph aggregator that is a kind of GCN. In that case, the filter relationship both in one layer and in the whole network can be automatically found. To the author’s knowledge, it is the first time that the graph convolution \cite{b14} is applied to model compression.
\item Compared with other AutoML methods, our method is more professional in pruning networks with short-cuts structure, such as ResNet serials and MobileNet-V2. Because a short-cut structure contains many loops in a graph whose features can be easily extracted by the graph aggregator.
\end{itemize}
\begin{figure*}
\centering
\includegraphics[width=5in]{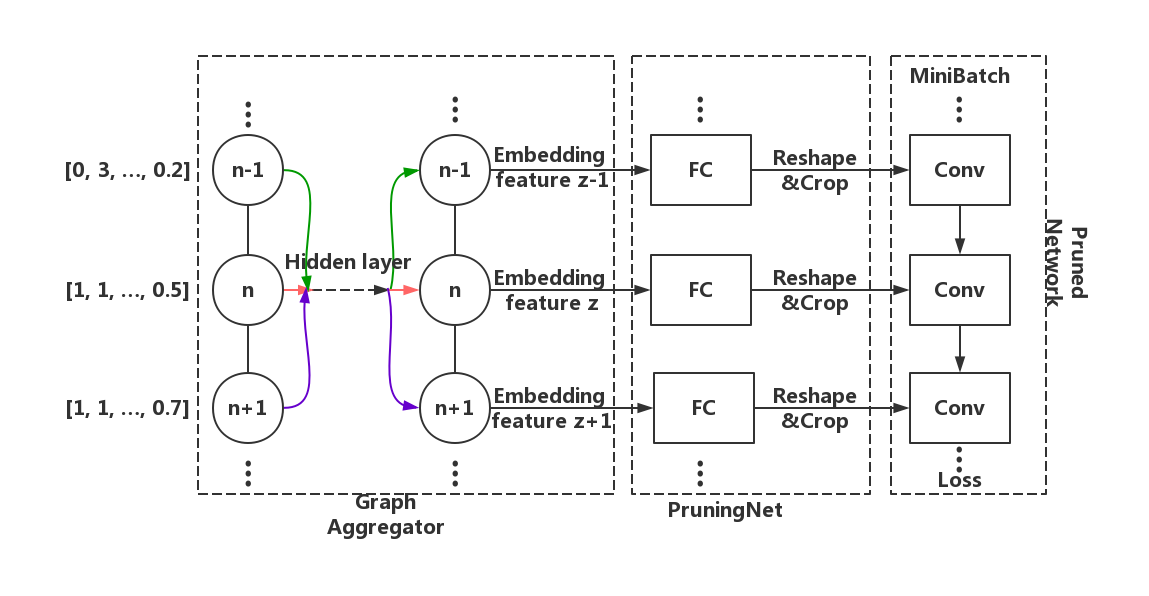}
\caption{The training chartflow. The Graph Aggregator is connected with Pruned Network by a PruningNet. The PruningNet consists of several fully connected(FC) layers. The weights of the Pruned Network is provided from the PruningNet by reshaping and crop its outputs to corresponding Conv weights shape.}
\label{fig:training}
\end{figure*}
\section{Related Works}
\textbf{Rule-based Channel Pruning} The target of channel pruning methods\cite{b16,b17,b18,b51,b52} is to accelerate the inference of large neural networks by reducing the number of channels, simultaneously, keep high accuracy. The key to successful channel pruning is to measure the importance of channels, i.e., $l_2$-norm of channel weights\cite{b17}, learnable scaled weights of batchnorm layer\cite{b16} and geometric median of channel weights\cite{b6}. Though the above rule-based methods have achieved remarkable improvement for CNN compression, it needs human-designed heuristics to guide pruning. However, manually defined rules are always sub-optimal for different tasks\cite{b8}. Besides, the above rule-based methods assumes different layers to be equally important\cite{b45}. As a matter of fact, channel weights between neighboring layers have a strong correlation so that the assumption in prior work is invalid. \\
\textbf{Graph Convolution Application} In the real world, many applications need to process non-Euclidean data, which cannot be effectively and thoroughly dealt with by normal CNNs. To extract the useful features from non-Euclidean data, many graph convolution networks(GCNs) are proposed to provide well-suited solutions for non-Euclidean data processing. The reliability and effectiveness of GCNs attract greatly interest in using GCNs for a variety of applications, such as social networks \cite{b21}, model chemical molecule structures \cite{b22,b23}, recommendation engines \cite{b24,b25} and natural language processing \cite{b26,b27}. Until now, graph convolution is still not applied in network structure analysis. Modern deep CNNs with short-cut structures often contain complex edges. It can be viewed as a connected graph so that GCNs can be used to extract information from the entire network structure. \\
\textbf{AutoML Pruning for Model Compression} Recently, AutoML pruning methods \cite{b8,b9,b10} have attracted a growing interest in automatically pruning for deep CNNs. Compared with pruning methods based on the human-craft rule, AutoML pruning methods aim to search for the best configuration without manual tuning. Our proposed GraphPruning also involves little human participation. Different from previous AutoML pruning methods, which only consider one layer of information, a graph aggregator is used to extract neighboring information for each node through converting the whole pruned network into a topology graph. Compared to previous AutoML pruning methods \cite{b8,b9}, GraphPruning can easily consider neighboring information in pruning and be compatible with other AutoML pruning methods.
\section{Methodology}
In this section, we first analyze the correlation between filters from different layers and take the 4th stage in ResNet-50 as an example. It is found that filters from the neighboring layer have a strong correlation, which means that the neighboring information should be taken into consideration in pruning.  \\
To solve this problem, the graph transformation method is presented which can easily describe any CNN network as a topology. For each node in the graph, some features are defined as embedding node features. These features are dynamically changed in the process of training and searching with different compression ratios. Further, a graph aggregator which is based on the graph convolution network is introduced to automatically pruning filters in CNN networks.\\
\subsection{Analysis of Filter Correlation} 
For many modern CNNs, the short-cut structure is indispensable. A block containing the short-cut structure can be formulated as 
\begin{equation}
\label{eq:shortcut}
y = x + F(x,{W_i}) 
\end{equation}
where $x$ and $y$ represent the input and output of this block respectively, $F$ denotes all of the convolution operations in this block and $W_i$ is the wights of $i$th convolution. \\
For the deep layer, deep networks can be converted into a shallow network by identity mapping, which means that eventually $F(x, {W_i}) = 0$ \cite{b3}. Although the short-cut structure accelerates the training of deep networks, it causes redundancy in the deep layer. To analyze filter correlation from different layers, a Pearson correlation matrix is designed. Pearson correlation matrix $P \in R^{m\times n}$ is generated by the features from two layers. Let two of the selected layers generate the feature maps $F^1 \in R^{h\times w\times m}$ and $F^2 \in R^{h\times w\times m}$, where $h, w,$ and  $m$ represent the height, width and the number of channels, respectively. The Pearson correlation matrix between the $i$th and the $j$th layer is formulated as
\begin{equation}
\label{eq:pearson}
\begin{split}
&P_{i,j}(x;W)\\
&= \sum\limits_{s= 1}^h  \sum\limits_{t= 1}^w |\frac {{F^1}_{s,t,i} (x^2;W^1) \times {F^2}_{s,t,j} (x^2;W^2)} { \sigma({{F^1}_{s,t,i} (x^1;W^1))} \times \sigma({{F^2}_{s,t,j} (x^2;W^2))} }|
\end{split}
\end{equation}
where $x^1$, $x^2$ and $W^1$, $W^2$ denote the input feature maps and weights of layer $1$ and layer $2$, respectively, $\sigma(\cdot)$ represents standard deviation. The $4$th stage from ResNet-50 \cite{b3} is taken as an example. As shown in \textbf{Fig. \ref{fig:correlation}}, As shown in Fig. 1, it is found that the convolution layer with kernel size $3\times 3$ has a strong correlation with its neighboring layer. Many filters in one layer can be represented by the related filters in other layers. Hence, the neighboring information should be fully considered in pruning. 
\subsection{Graph Transformation} 
In this subsection, a method that can easily convert a network to a graph is proposed while the way how to transform a neural network into a topology graph is first introduced. Shown as \textbf{Fig. \ref{fig:topology}}, the central idea of graph transformation is as follows: for convolution layer $L_i$ , it can be viewed as node $n_i$ in the graph $G$, and edge $e_{i,j}$ that exists if layer $L_i$ and $L_j$ are connected directly. For each node, it contains 7 features that characterize the state of the node:
\begin{equation}
\label{wq:feature}
\begin{split}
&(type, in\ channels, out\ channels, stride, \\ &kernel, weight\ size, ratio)
\end{split}
\end{equation}
\textbf{type:} the type of operation for each node. For simplicity, batch normalization \cite{b36} and nonlinear (ReLu) \cite{b40} operation are ignored in the graph so that the types of operation for modern CNNs are classified as normal convolution, depth-wise convolution and addition.\\
\textbf{in channels:} the number of feature maps inputted to each node. It will be changed in training and searching by multiplying the ratio of this node.\\
\textbf{out channels:} the number of feature maps outputted by each node. It will be changed in training and searching by multiplying the ratio of the former node. \\
\textbf{stride:} the stride of convolution operation. It will be set to 0 if the type of the operation is not convolution. \\
\textbf{kernel:} the kernel size of the convolution operation. It will be set to 1 if the type of operation is not convolution. \\
\textbf{weight size:} the product of the parameter dimension of each node. It will be set to 0 for the add and concatenation type. In training and searing, it will be changed by the multiplying ratio of this node and the former node.\\
\textbf{ratio:} the compression ratio for each node. This feature will be given by a random uniform in the process of training. In the process of searching, it will be given by a DDPG agent.\\
For this graph $G$, its elements $a_{i,j}$ of the adjacency matrix $A$ can be defined by
\begin{equation}
\label{eq:adj_element}
a_{i, j} =  \begin{cases}1 & if\ there\ exists\ e_{i,j} \\0 & other\end{cases} 
\end{equation}
To alleviate exploding/vanishing gradients \cite{b29} in GCN training, adjacency matrix $A$ will be renormalized by
\begin{equation}
\label{eq:a}
\hat{A} =  \tilde{D}^{1/2}  \tilde{A}  \tilde{D}^{-1/2}
\end{equation}
where $\tilde{A} = A + I_N$ and $I_N$ is the identity matrix with N dimension.
\begin{figure}
\centering
\includegraphics[height=3.5cm,width=8cm]{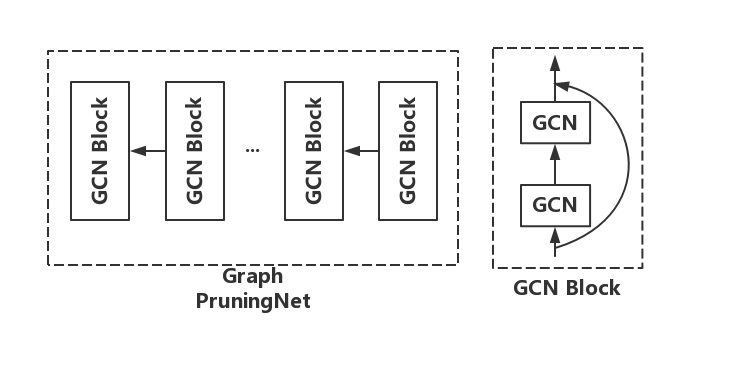}
\caption{The graph aggregator consists of some GCN blocks. Each GCN block contains two GCN layers and a short-cut structure to avoid over smooth \cite{b28}.}
\label{fig:GCN}
\end{figure}
\subsection{Graph PruningNet }
To gather the neighbor features for each node, a graph aggregator based on GCN is constructed. The structure of the graph aggregator is given in \textbf{Fig. \ref{fig:GCN}}. Residual GCN architecture \cite{b28} with some blocks which contain 2 graph convolution layers is used. The forward of each block can be simply described as
\begin{equation}
\label{eq:gcn_forward}
Z = X + ReLU(\hat{A}ReLU(\hat{A}XW^{(0)})W^{(1)})
\end{equation}
where $W^{(0)} \in R^{H \times C}$, $W^{(1)} \in R^{C \times F}$ are the parameters of the first and second graph convolution layer, respectively. $Z \in R^{F \times N}$ is the high-level feature map of the graph with $N$ nodes. $X$ denotes the inputs of the block. The nonlinear function ReLU, defined as $max(0, x)$ with input $x$, is used.\\
Inspired by the MetaPruning \cite{b9}, the graph aggregator and the Pruned Networks are connected by a PruningNet so that the graph aggregator can be trained jointly with the PruingNet on the dataset, meanwhile, the PruningNet can provide reasonable weights for pruned layer. The difference between our approach and MetaPruning is that node features are utilized as inputs of the PruningNet rather than the human-defined individual layer features.\\
The embedding matrix of graph $C=(c_1, c_2, ..., c_l)$, where $c_i \in R^{1\times 7}$ denotes the embedding feature (3) of $i$th node, will be taken as input for the graph aggregator $G$ with its weights $\theta_G$. $N=(n_1, n_2, ..., n_l)$ is the aggregated embeddingg matrix. Then the $i$th FC layer with its weights $\theta_i$ takes the embedding vector $n_i$, where $n_i$ denotes the $i$th column of output with the graph aggregator, as input to generate the weights $W_i$ of pruned layer 
\begin{equation}
\label{eq:weights}
\begin{split}
&(n_1, n_2, ..., n_l) = G(c_1, c_2, ..., c_l|\theta_G)\\
&W_i = FC_i(n_i|\theta_i)
\end{split}
\end{equation}
\textbf{Stochastic training} The training chartflow is shown in \textbf{Fig. \ref{fig:training}}. In each training step, the compression ratio $r_i$ of $i$th pruned layer is randomly given by Uniform distribution, while the corresponding node features $c_i$ in the graph also is changed according to $r_i$. According to \textbf{Eq.(\ref{eq:weights})}, the reasonable weights $W=(W_1, W_2, ..., W_l)$ of the Pruned Network are generated. Then the Pruned Network is trained on the dataset. The whole forward can be given by
\begin{figure*}
\centering
\includegraphics[width=5in]{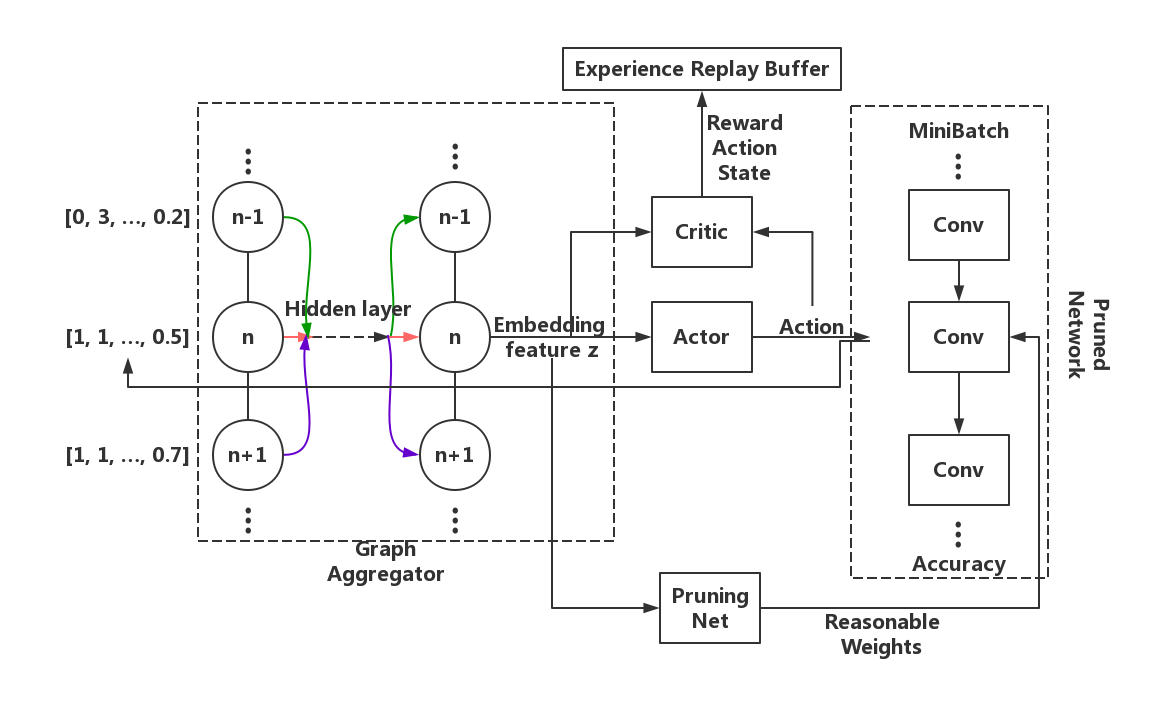}
\caption{In searching, our DDPG agent receives embedding features from node $n_i$ and outputs a compression ratio of $a_t$. The weights of Pruned Networks is given by trained PruningNet.}
\label{fig:searching}
\end{figure*}
\begin{equation}
\label{eq:forward}
\begin{split}
&Y = F(X, W)\\
&L = CrossEntropy(Y, Y')
\end{split}
\end{equation}
where $Y$ and $Y'$ are the inference result of the Pruned Network and the label of $X$ respectively, $L$ denotes the CrossEntropy loss. After inference, back-propagated gradients are calculated for weights of the graph aggregator and FC layers according to chain rule as follow
\begin{equation}
\label{eq:gradient}
\begin{split}
&\nabla \theta_G=\frac{\partial L}{Y} \frac{\partial Y}{W} \frac{\partial W}{N} \frac{\partial N}{\theta_G}\\
&\nabla \theta_i=\frac{\partial L}{Y} \frac{\partial Y}{W_i} \frac{\partial W_i}{\theta_i}
\end{split}
\end{equation}
In this way, the graph aggregator can learn to extract high-level embedding features for each node while each FC layer can learn how to generate reasonable weights for each pruned layer end to end. The detailed algorithm is described in \textbf{Alg. \ref{alg:train}}.\\
\begin{algorithm}[t]\small
\caption{Training Algorithm for GraphPruning}
\label{alg:train}
\begin{algorithmic} 
\REQUIRE graph aggregator $G$, pruning net $FC$, pruned network $F$, iteration number $N$
\ENSURE trained graph aggregator $G$, trained pruning net $FC$
\STATE Construct a graph for pruned network and initiallize node features $C$
\FOR{$i=1$ to $N$} 
\STATE Randomly sample ratio $r$;
\STATE Update node features $C$ according to $r$;
\STATE Generate weights for pruned network by Eq.(\ref{eq:weights});
\STATE Get a batch from data and forward to get $L$ by Eq.(\ref{eq:forward});
\STATE Backward $L$ to both $\theta_G$ and $\theta_i$ by Eq.(\ref{eq:gradient});
\STATE Update graph aggregator $G$ and pruning net $FC$;
\ENDFOR
\end{algorithmic}
\end{algorithm}  
\begin{algorithm}[t]\small
\caption{Searching Algorithm for GraphPruning}
\label{alg:search}
\begin{algorithmic} 
\REQUIRE trained graph aggregator $G$ from Alg.\ref{alg:train}, trained pruning net $FC$ from Alg.\ref{alg:train}, pruned network $F$, layer number $L$
\ENSURE the optimal pruning ratio $r_b$
\STATE Construct a graph for pruned network and initiallize node features $C$
\WHILE{not converged} 
\FOR{$i=1$ to $L$}
\STATE Generate the current state $s_i$ by Eq.(\ref{eq:reward_state});
\STATE Sample prune ratio $r_i$ by Eq.(\ref{eq:action});
\STATE Prune current layer;
\ENDFOR
\STATE Obtain reward $R$ by evaluating the pruned network;
\STATE Save each transition $(s_i, a_i, R, s_{i+1})$ to experience replay buffer;
\STATE Get a batch from experience replay buffer;
\STATE Update actor and critic by Eq.(\ref{eq:bellman});
\ENDWHILE
\end{algorithmic}
\end{algorithm}  
\textbf{DDPG searching} The searching chartflow is revealed in \textbf{Fig. \ref{fig:searching}}. After training the graph aggregator and each FC layer, the accuracy of trained Pruned Networks with any channel configuration can be easily obtained. Thus, the weights of the graph aggregator and FCs are not updated in searching. However, the configuration that can construct the best Pruned Network under computational constraints is still needed to be searched. The DDPG algorithm which is an off-policy actor-critic algorithm is used to search over ratio space. The reason why the reinforcement learning method is used rather than the evolutionary method is that: (1) In the training phase, the graph aggregator has been trained well, which can gather neighbor features for each node. The features of each node can be viewed as the states in reinforcement learning. (2) Pruning each layer is a sequence decision, especially since neighbor information is taken into consideration. \\
The agent prunes the model layer by layer. For each layer, the actor takes corresponding node features as the state and outputs an action $a\in [0,1]$ where the action denotes the compression ratio. Following previous AutoML works with reinforcement learning \cite{b8,b38,b39,b40}, each transition in an episode is $(s_i, a_i, r, s_{i+1})$ which will be saved in the experience replay buffer where the reward $R$ and the state $s_i$ can be formulated as:
\begin{equation}
\label{eq:reward_state}
\begin{split}
&R = accuracy \\
&s_i = G(c_1, c_2, ..., c_l)[i]\\
\end{split}
\end{equation}
During searching, the actor-network generates the pruning ratio for each layer. For the exploration noise process, the truncated normal distribution is used:
\begin{equation}
\label{eq:action}
r_i \sim{~} TN(\mu(s_i), \eta^2, 0, 1)
\end{equation}
where $\mu^{'}$ and $\mu$ denote the sampled action and probability distribution of action, respectively. Noise $\eta$ is decayed after each episode exponentially. During the update, a variant form of Bellman's Equation \cite{b13} is utilized:
\begin{equation}
\label{eq:bellman}
\begin{split}
&L_r = \frac{1}{N}\sum_{i=1}^{N}(y_i - Q(s_i, a_i))^2 \\
&y_i = \hat{R_i} + \gamma Q(s_{i+1}, \mu_{i+1})
\end{split}
\end{equation}
where $Q$ and $\hat{R}$ represent the critic network and normalized reward, respectively. The discount factor $\gamma$ is set to avoid over-prioritizing short-term rewards. We show the detail of the searching process in \textbf{Alg. \ref{alg:search}}.
\section{Experiments}
\begin{table*}
\caption{Experimental configurations.} 
\label{exp_config}
\begin{center}
\begin{tabular}[c]{c|c|c|c|c|c|c}
\hline
Phase&Model&Epochs&Batch size& Init LR& Weight decay& Data augmentation\\
\hline
&MobileNetV1&64&512&0.25&$4\times10^{-5}$&crop + flip\\
Train&MobileNetV2&64&512&0.25&$0$&crop + flip\\
&ResNet-50&32&256&0.1&$1\times10^{-4}$&crop + flip + lighting\\
\hline
&MobileNetV1&256&1024&0.5&$4\times10^{-5}$&crop + flip + lighting\\
Retrain&MobileNetV2&256&1024&0.5&$4\times10^{-5}$&crop + flip + lighting\\
&ResNet-50&128&512&0.2&$1\times10^{-4}$&crop + flip + lighting\\
\hline
\end{tabular}
\end{center}
\end{table*}
The graph aggregator is an expert in extracting features from the topology graph with many loops. Hence, the effectiveness of our approach is demonstrated by pruning MobileNet V1 \cite{b1}, MobileNet V2 \cite{b2} and ResNet-50 \cite{b3}. Besides, as the same as MetaPruning \cite{b9}, the iterative fine-tuning process in evaluation is not involved in our approach so that all experiments can be easily carried out on a large-scale dataset, ImageNet-2012 \cite{b15}, which has 1.28 million training images and 50k validation images of 1000 classes. We use top-1 accuracy to measure the performance of models in our experiments since ImageNet-2012 has balanced samples for each class. Deep ROC analysis\cite{b47} can be applied to measure the performance of models in the case of the dataset with unbalanced samples.
\subsection{Experimental Settings}
A graph aggregator with 2 GCN blocks is constructed. The number of hidden layers for each GCN block is set to 64. The whole process of pruning can be split into two stages. First, we jointly train the graph aggregator, FC layers and the Pruned Network on the training dataset. After training, the well-trained graph aggregator will be used to produce a state for the DDPG agent. \\
\textbf{Training setting} On ImageNet-2012, we randomly split the original training data into two parts, the sub-validation dataset contains 50000 images and the sub-training dataset has the rest of the images. Data augmentation strategies following PyTorch \cite{b34} official examples are applied. For a fair comparison with \cite{b9}, the training scheme is followed as \cite{b9} both on ResNet-50 \cite{b3} and MobileNet \cite{b1,b2}. The scheme detail of each model is shown in \textbf{Table.\ref{exp_config}}. Because of stochastic training, it causes the problem of feature aggregation inconsistency \cite{b35} so that moving average statistics of means and variances in training image which is computed in Batch Normalization \cite{b36} cannot be as means and variances in the test image. Thus, as it is shown on\cite{b10}, all BatchNorm layers are privatized for different channel widths. \\
\textbf{Searching setting} At first, the actor and critic are constructed, which simply consists of two FC layers, for the DDPG agent. To conduct a fair comparison with AMC \cite{b8}, the reinforcement learning settings are followed as \cite{b8}. Noise initialized as 0.5 and decayed after each episode exponentially is used to enhance exploration. In each searching step, we fix the ratio for all pruned layers and recalibrate the moving average statistics for all Batch Normalization layers on recalibration data. \\
\textbf{Retraining setting} After training and searching, the best configuration and corresponding weights are taken to construct a new model. Next, the new model is trained from scratch with the same scheme as \cite{b9}. The scheme detail of each model is shown in \textbf{Table.\ref{exp_config}}.
\subsection{MobileNet Pruning}
\textbf{MobileNet V1} A single-branch network, MobileNet V1, is pruned by the proposed approach. The topology graph of MobileNet V1 is quite simple, which consists of 27 nodes and 26 edges.\\ 
\begin{figure*}[htbp]
\subfigure[]{
\begin{minipage}[t]{0.45\linewidth}
\centering
\includegraphics[width=1.5in]{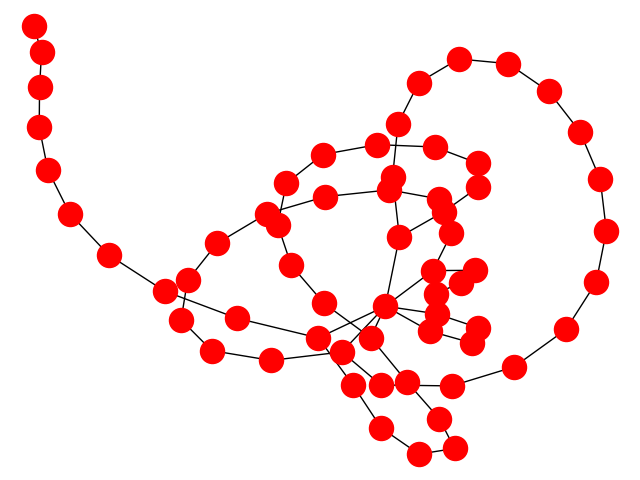}
\end{minipage}%
}%
\subfigure[]{
\begin{minipage}[t]{0.45\linewidth}
\includegraphics[width=1.5in]{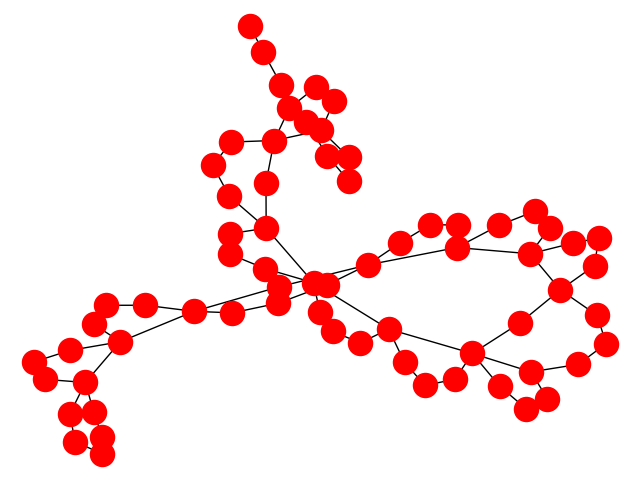}
\end{minipage}%
}%
\centering
\caption{(a) topology graph converted from MobileNet V2. (b) topology graph converted from ResNet-50.}
\label{fig:topology_network}
\end{figure*}
In each training batch, a ratio $r\in [0, 1]$ is randomly set for each node and the graph feature map $M \in R^{27\times 7}$ will be changed by corresponding ratio $r$. Then, the graph aggregator generates a higher dimensional graph feature map $H \in R^{27\times 64}$. The $i$th FC layer takes embedding features $H[i]$ as input and generates a vector $L \in R^{1\times C_{out} \times C_{in} \times K^2}$ where $C_{out}$, $C_{in}$ and $K$ denote the number of out channels and in channels, the kernel size of the $i$th Pruned layer. At last, we crop and reshape $L$ to $(C_{out} \times r_{i}, C_{in} \times r_{i-1}, K, K)$ as the reasonable weights for Pruned layer.\\
During the searching procedure, we also randomly set the ratio $r$ for each node at the first 100 episodes to warm up. After warming up, the DDPG agent takes $H[i]$ and outputs a ratio $r$ to prune the $i$th layer, meanwhile, graph feature map $M$ is dynamically changed with the ratio $r$. It is noted that only the point-wise convolutional layers are pruned and the ratio of depth-wise convolutional layers will be set as the same as the ratio of their former convolution.\\
\textbf{MobileNet V2} We prune MobileNet V2, a highly compact network consisting of depth-wise convolution and point-wise convolution layers. As it is shown on \textbf{Fig. \ref{fig:topology_network} (a)}, a topology graph that contains 62 nodes and 71 edges is used to represent the MobileNet V2. The process of training and searching is like pruning on MobileNet V1. Different from MobileNet V1, MobileNet V2 has a short-cut structure for each block that does not have downsampling. Different with \cite{b6,b17}, we also prune the projection shortcuts by sharing ratio $r$ for pruned layer where output is directly connected with a shortcut. \\ 
We compare our method with some state-of-the-art network pruning methods. The results of the comparison on MobileNet V1 and V2 are shown in \textbf{Table \ref{table:imagenet}}. For MobileNet V1, GraphPruning achieves the same top-1 accuracy with AMC under FLOPs 285M, but our method achieves the state-of-the-art which boosts \textbf{0.3\%} top-1 accuracy with MetaPruning when the model size is as small as 41M FLOPs. For MobileNet V2, Because it contains many short-cut structures from which information can be extracted by the graph aggregator, our method gains improvement over MetaPruning by 0.6\% under 43M FLOPs.\\
To evaluate the realistic acceleration, the forward time of the Pruned Network is measured on one 2080Ti GPU. The results of MobileNet V1/V2 are shown in \textbf{Table \ref{table:mobilenetv1}}/\textbf{Table \ref{table:mobilenetv2}}. Under the same compression ratio, the method keeps close or better inference time but obtains better top-1 accuracy.
\begin{table*}[!t]
\caption{Results of ImageNet classification. We show the top-1 accuracy of each method under the same or closed FLOPs. }
\label{table:imagenet}
\begin{center}
\begin{tabular}[c]{c|ccc}
\hline
Pruned Network&Method&top-1 
acc.(\%)&FLOPs\\
\hline
&Baseline &68.4&325M\\
&AMC &70.5&285M\\
&Slimmable Network &69.5&325M\\
&MetaPruning &70.4&281M\\
MobileNet V1&\textbf{GraphPruning}&\textbf{70.5}&285M\\
\cline{2-4}
&Baseline &50.6&41M\\
&MetaPruning&57.2&41M\\
&Slimmable Network&53.1&41M\\
&\textbf{GraphPruning}&\textbf{57.5}&41M\\
\hline
\hline
&Baseline&69.8&220M\\
&AMC &70.8&220M\\
&MetaPruning &71.2&220M\\
&Slimmable Network&68.9&209M\\
MobileNet V2&GraphPruning&71.6&220M\\
\cline{2-4}
&Baseline &54.3&43M\\
&MetaPruning &58.3&43M\\
&\textbf{GraphPruning}&\textbf{58.9}&43M\\
\hline
\hline
&Baseline 1.0$\times$&76.6&4.1G\\
&Baseline 0.75$\times$ &74.8&2.3G\\
&Slimmable Network &74.9&2.3G\\
&MetaPruning&75.4&2.3G\\
&AOFP-C1 \cite{b43}&75.63&2.58G\\
&C-SGD-50 \cite{b42}&74.54&1.7G\\
ResNet-50&ThiNet-50 \cite{b41}&74.7&2.1G\\
&\textbf{GraphPruning}&\textbf{76.1}&2.3G\\
\cline{2-4}
&Baseline 0.5$\times$&72.0&1.0\\
&Slimmable Network&72.5&1.0G\\
&ThiNet-30 \cite{b41}&72.1&1.2G\\
&MetaPruning &73.4&1.G\\
&\textbf{GraphPruning}&\textbf{74.7}&1.0G\\
\hline
\end{tabular}
\end{center}
\end{table*}
\begin{table*}
\caption{Comparison of the realistic inference time with MobileNet V1.} 
\label{table:mobilenetv1}
\begin{center}
\begin{tabular}[c]{c|c|c|c|c}
\hline
Ratio&Baseline time(ms)&Pruned time(ms)&Baseline Acc.(\%)&Pruned Acc.(\%)\\
\hline
$1\times$&0.65&-&70.9&-\\
$0.75\times$&0.51&0.51&68.4&69.9\\
$0.5\times$&0.35&0.35&63.3&67.7\\
$0.25\times$&0.21&0.20&49.8&59.4\\
\hline
\end{tabular}
\end{center}
\end{table*}
\begin{table*}
\caption{Comparison of the realistic inference time with MobileNet V2.} 
\label{table:mobilenetv2}
\begin{center}
\begin{tabular}[c]{c|c|c|c|c}
\hline
Ratio&Baseline time(ms)&Pruned time(ms)&Baseline Acc.(\%)&Pruned Acc.(\%)\\
\hline
$1\times$&0.72&-&71.7&-\\
$0.75\times$&0.53&0.52&69.8&71.0\\
$0.5\times$&0.41&0.41&65.4&68.9\\
$0.35\times$&0.33&0.32&60.3&64.2\\
\hline
\end{tabular}
\end{center}
\end{table*}
\subsection{ResNet-50 Pruning}
ResNet-50 consists of four stages which are stacked by some blocks which consists of a $1\times 1$ convolution, a $3\times 3$ convolution, a $1\times 1 covolution$ and an extra $1\times 1$ convolution when $stride = 2$. It is shown that its topology graph in \textbf{Fig. \ref{fig:topology_network} (b)}. In the analysis process of \textbf{subsection 3.1}, due to its short-cut structure in Bottlenecks, many filters in $3\times 3$ convolution are redundant and can be represented by filters in other convolution. Thus, the projection shortcuts are pruned as pruned on MobileNet V2. Different from MobileNet V2, the block with $stride = 2$ in ResNet-50 has an extra $1\times 1$ convolution in the residual path. To keep the number of channels the same for adding operation at the last of the block, the one ratio for the extra convolution and the last convolution are shared. \\
We test GraphPruning on ResNet-50 with pruning rates of 0.5 and 0.25. The comparison of results with other AutoML methods is revealed in \textbf{Table \ref{table:imagenet}} and GraphPruning outperforms previous methods again.\\ 
\subsection{Results Visualization and Discussion}
\begin{figure*}
\subfigure[]{
\begin{minipage}[t]{0.33\linewidth}
\centering
\includegraphics[width=1.8in]{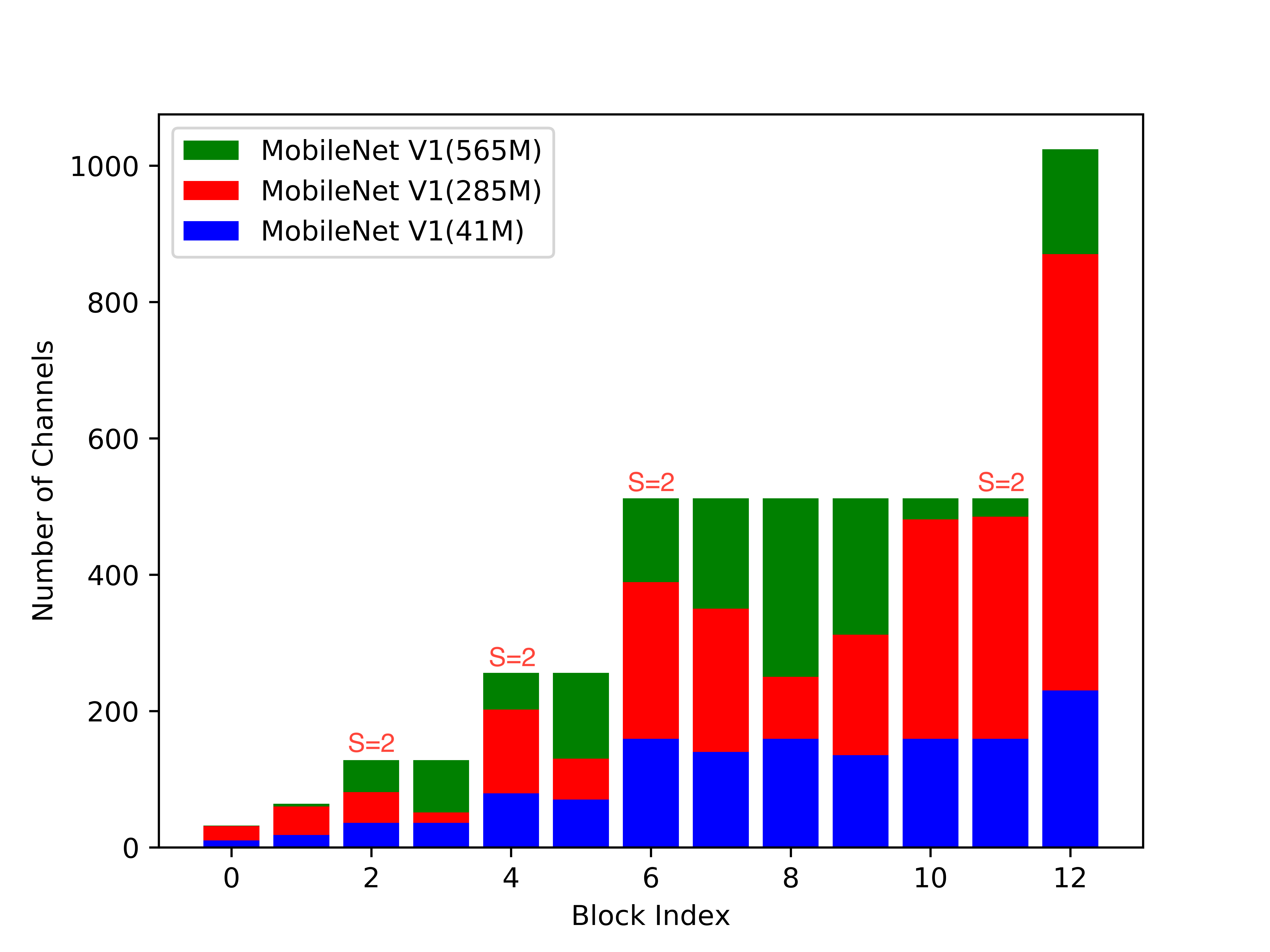}
\end{minipage}%
}%
\subfigure[]{
\begin{minipage}[t]{0.33\linewidth}
\includegraphics[width=1.8in]{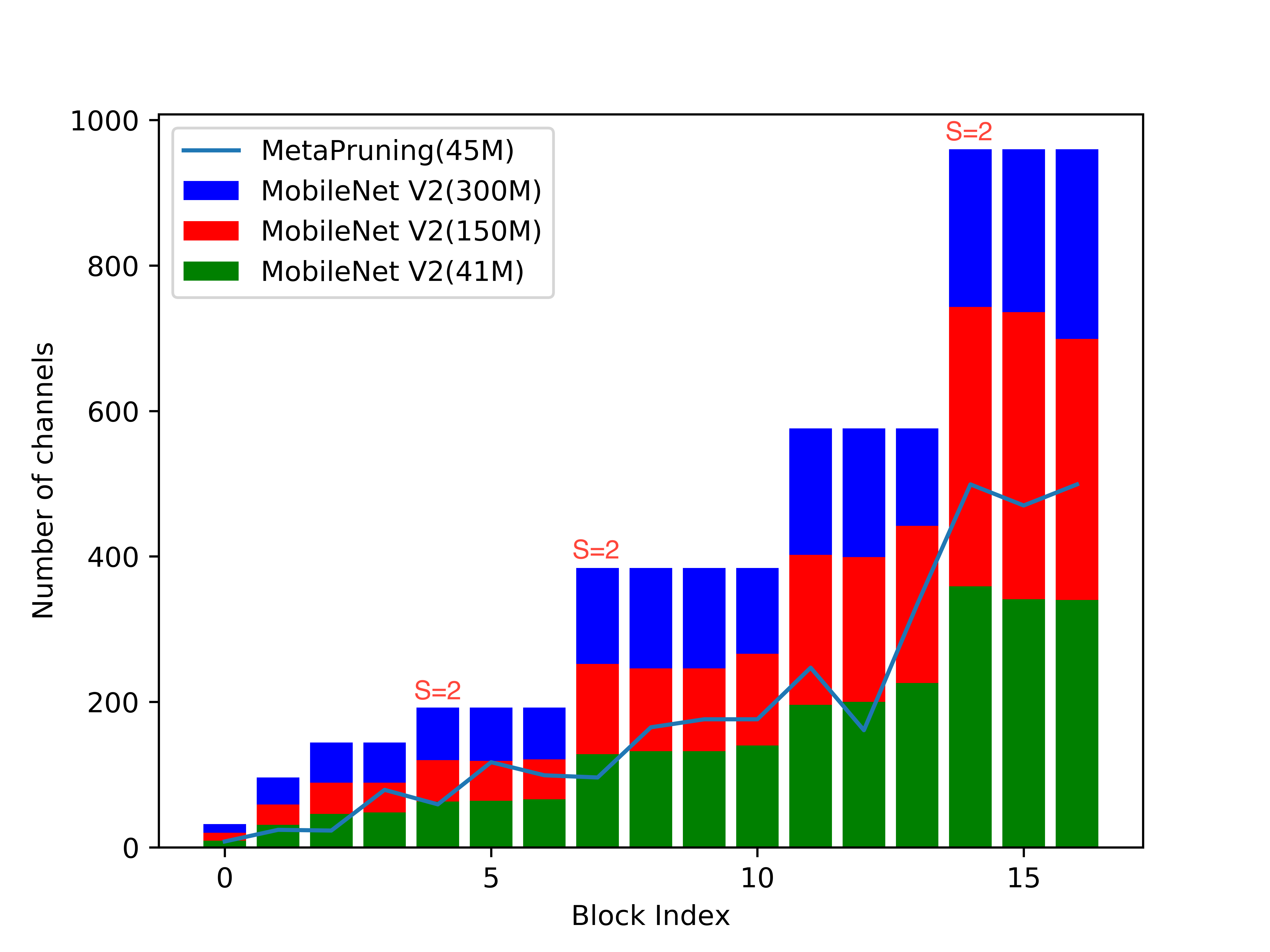}
\end{minipage}%
}%
\subfigure[]{
\begin{minipage}[t]{0.33\linewidth}
\includegraphics[width=1.8in]{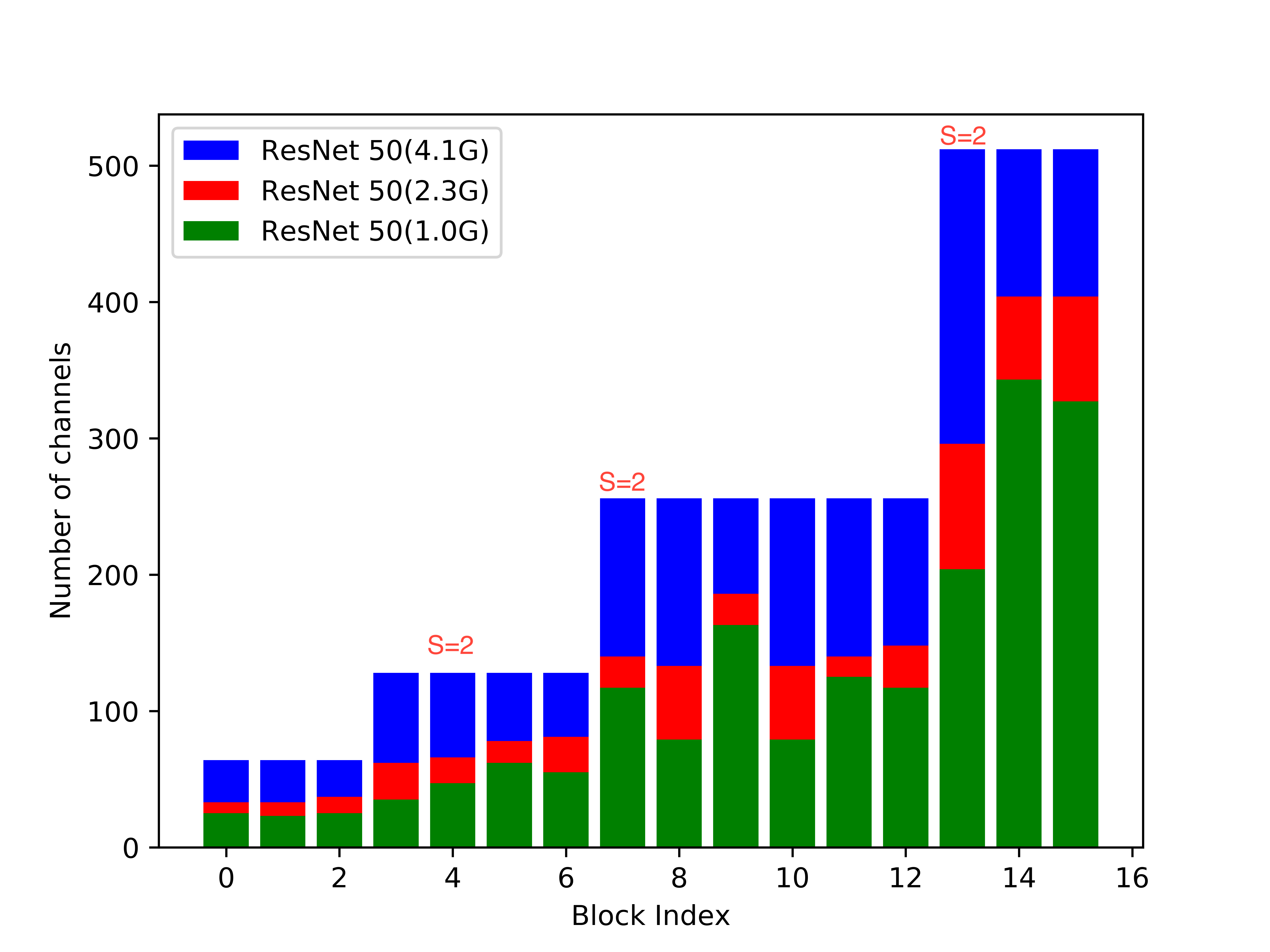}
\end{minipage}%
}%
\centering
\caption{This figure shows the number of output channels of each block in a different model. The line labeled with Graph pruning denotes our method under different FLOPs constraints. The layer with $stride=2$ is marked with a dashed box. (a) MobileNet V1 at 565M, 285M, and 41M FLOPs. (b)MobileNet V2 at 300M, 220M, 150M, 41M, 45M FLOPs. (c)ResNet-50 at 4.1G, 1.0G, and 2.3G FLOPs.}
\label{fig:arch}
\end{figure*}

\begin{figure*}
\subfigure[]{{\label{fig2(a)}}
\begin{minipage}[t]{0.3\linewidth}
\centering
\includegraphics[width=1.8in]{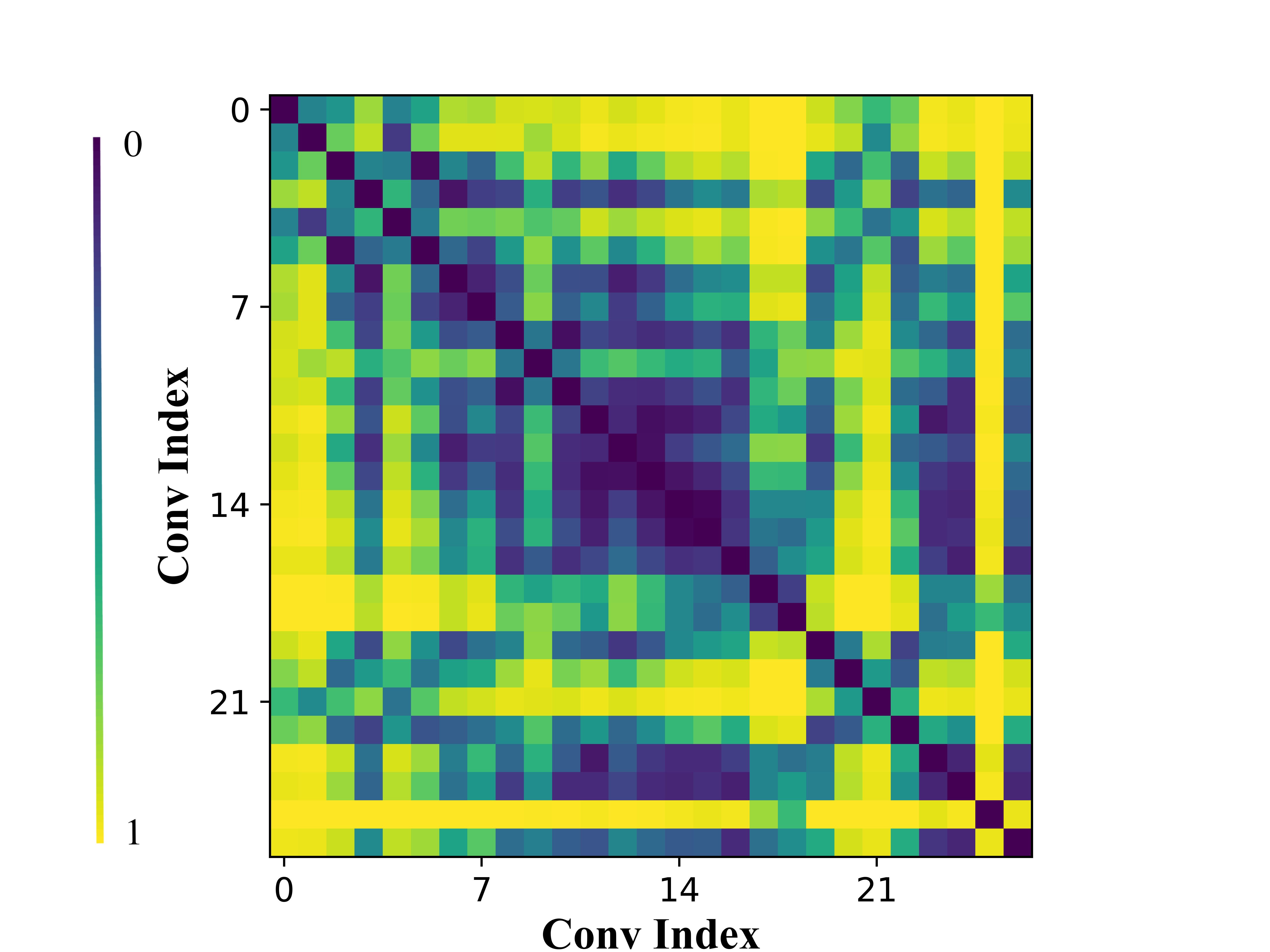}
\end{minipage}}
\subfigure[]{{\label{fig2(b)}}
\begin{minipage}[t]{0.3\linewidth}
\centering
\includegraphics[width=1.8in]{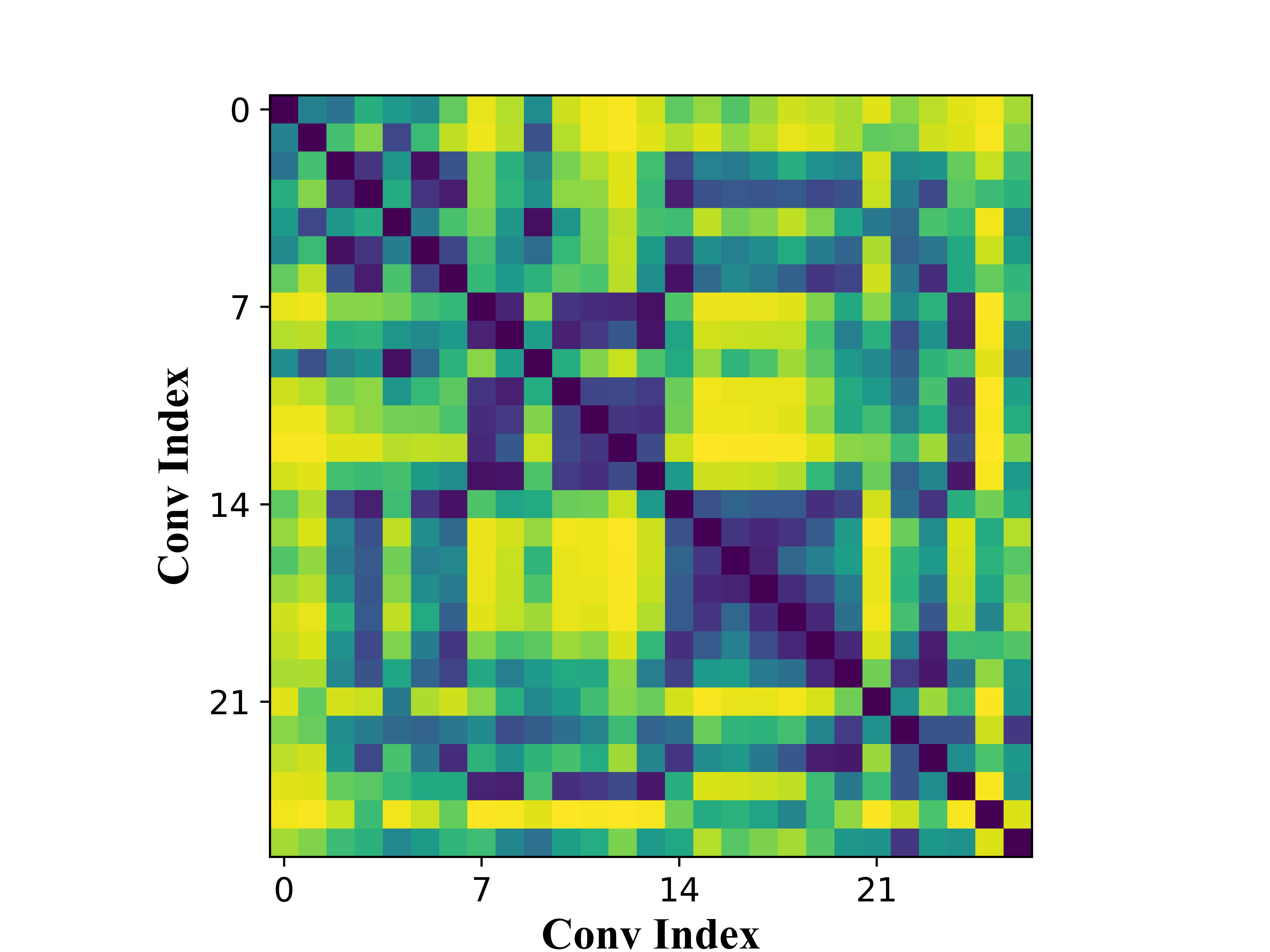}
\end{minipage}}
\subfigure[]{{\label{fig2(b)}}
\begin{minipage}[t]{0.3\linewidth}
\centering
\includegraphics[width=1.8in]{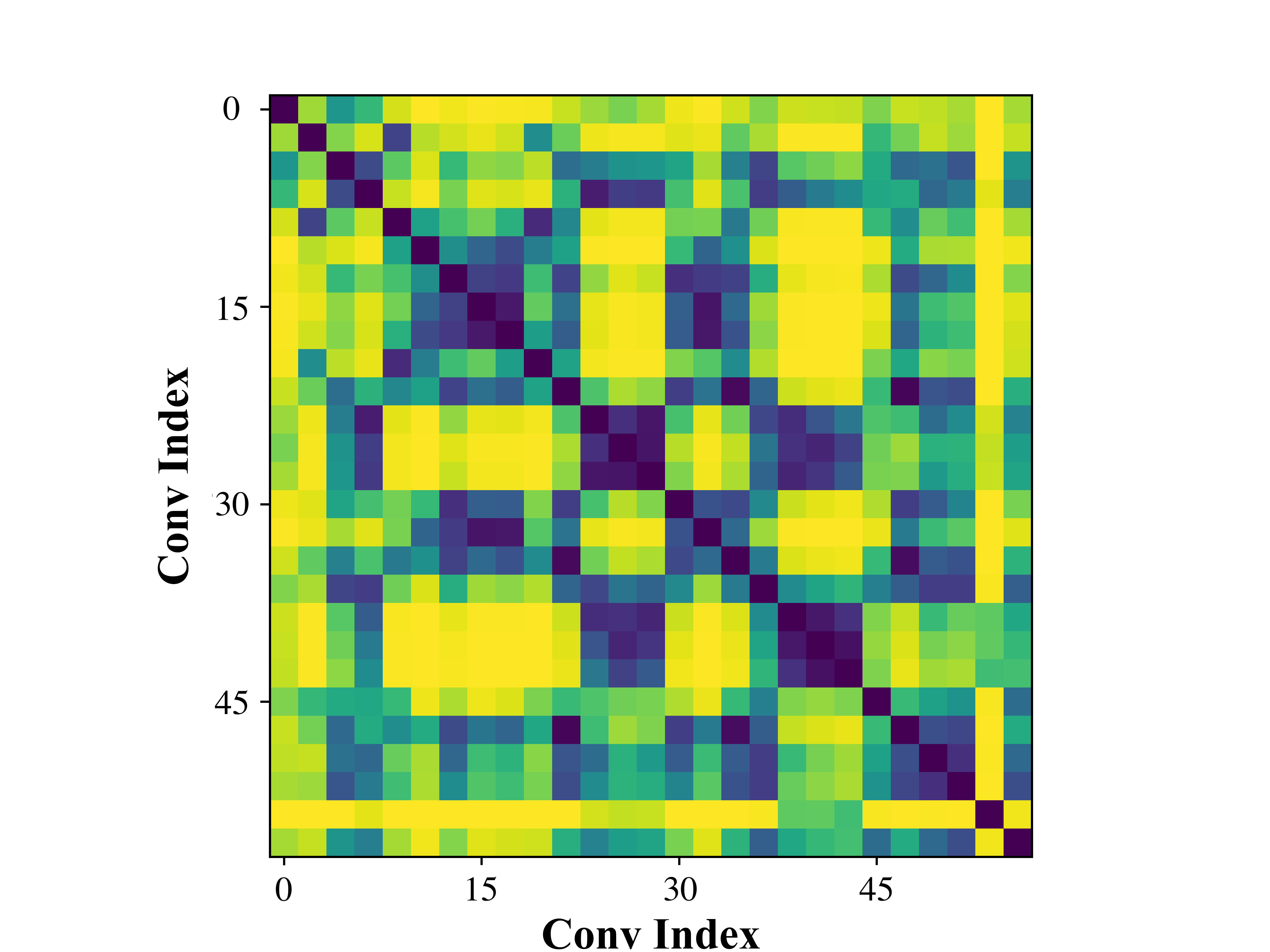}
\end{minipage}}
   \caption{These figures show the distance of features between each node after gathering the neighbor features. (a) MobileNet V1. (b) MobileNet V2. (c) ResNet-50.}
\label{fig:neighbor_feature}
\end{figure*}

We visualize what the graph aggregator learned by measuring the MSE of node features after gathering. As shown in \textbf{Fig \ref{fig:neighbor_feature}}., the distance between the neighbor convolutional layer is small, which means the features of each node have fully aggregated the neighbor features. \\
The best configuration of Pruned Networks is visualized to know more about what the DDPG agent has learned. \\
\textbf{MobileNet V1} The best configuration of MobileNet V1 is analyzed. the block in MobileNet V1 contains a $3\times 3$ depth-wise convolution and a $1\times 1$ point-wise convolution which has no short-cut structure. The best configuration of out channels for each block is shown in \textbf{Fig. \ref{fig:arch}(a)}. It is found that convolution with $stride = 2$ keeps more channels which have been mentioned in \cite{b9}. Besides, more channels are kept in deep layers when pruning less.\\
\textbf{MobileNet V2} A block from MobileNet V2 consists of a $1\times 1$ point-wise convolution, a $3\times 3$ depth-wise convolution and a $1\times 1$ point-wise convolution, which has a short-cut structure when $stride=1$. By extracting the number of channels from middle convolution in each block, as shown in \textbf{Fig. \ref{fig:arch}(b)}, a similar phenomenon is found that more channels are kept in convolution with $stride = 2$ and deep layer. Compared with \cite{b9}, our method keeps more channels when $stride=2$ and prunes more aggressively on other layers. During the analysis procedure in \textbf{subsection 3.1}, the short-cut structure causes redundancy. The block with $stride=2$ in MobileNet V2 does not have short cut yet so that it contains fewer redundant filters than others.\\
\textbf{ResNet-50} As same as the analysis of MobileNet V2, we also extract the number of channels from the middle convolution in each block. Different from the block in MobileNet V2, each block in ResNet-50 has a short-cut structure and contains an extra $1\times 1$ convolution when $stride = 2$. The best configurations under different FLOPs constraints are revealed in \textbf{Fig. \ref{fig:arch}(c)}, it is found that the agent has learned the policy that more channels are kept in the deep layer.\\
\textbf{Discussion} The regular patterns of pruning results for MobileNet V1/V2 and ResNet-50 are discussed here. \\
As it is shown on \textbf{Figs. \ref{fig:arch}}, all of these pruned networks keep more channels in the deep layer. This phenomenon also exists in many other AutoML pruning methods \cite{b8,b9}. Liu \cite{b9} suspected that it is caused by the number of classifiers for the ImageNet dataset which contains 1000 classes. However, in AMC \cite{b8}, more channels are also kept in deep layers when searching on CIFAR-10 \cite{b37} dataset which only contains 10 classes. Hence, we explain it from the perspective of the original Pruned structure. The resolution of the input picture is very large so that large channel numbers cannot be set in the shallow layer because of GPU memory constraints. To solve this problem, modern deep CNN models \cite{b2,b3,b4} are designed deep but narrow, which means the deep layer has more channels than the shallow layer. We cannot prune deep layers excessively to ensure a sufficient feature map for the whole network.
Besides, the policy that the agent has learned is more preferred to keep channels in layers with $stride = 2$. Convolution operation with $stride = 2$ will shrink the resolution of feature maps. The shrink in the size of the feature map leads to spatial information transform so that the filter correlation is also reduced. As Shown in \textbf{Fig. \ref{fig:correlation}.(a)} and \textbf{(b)}, the correlation of feature map is significantly reduced after the feature map is processed by convolution layer with $stride = 2$. Thus, it is reasonable that the policy keeps more channels in the layer with $stride = 2$. \\
\hspace*{1em}\subsection{Ablation Study} 
\begin{figure}
\subfigure[]{
\begin{minipage}[t]{0.5\linewidth}
\centering
\includegraphics[width=2.4in]{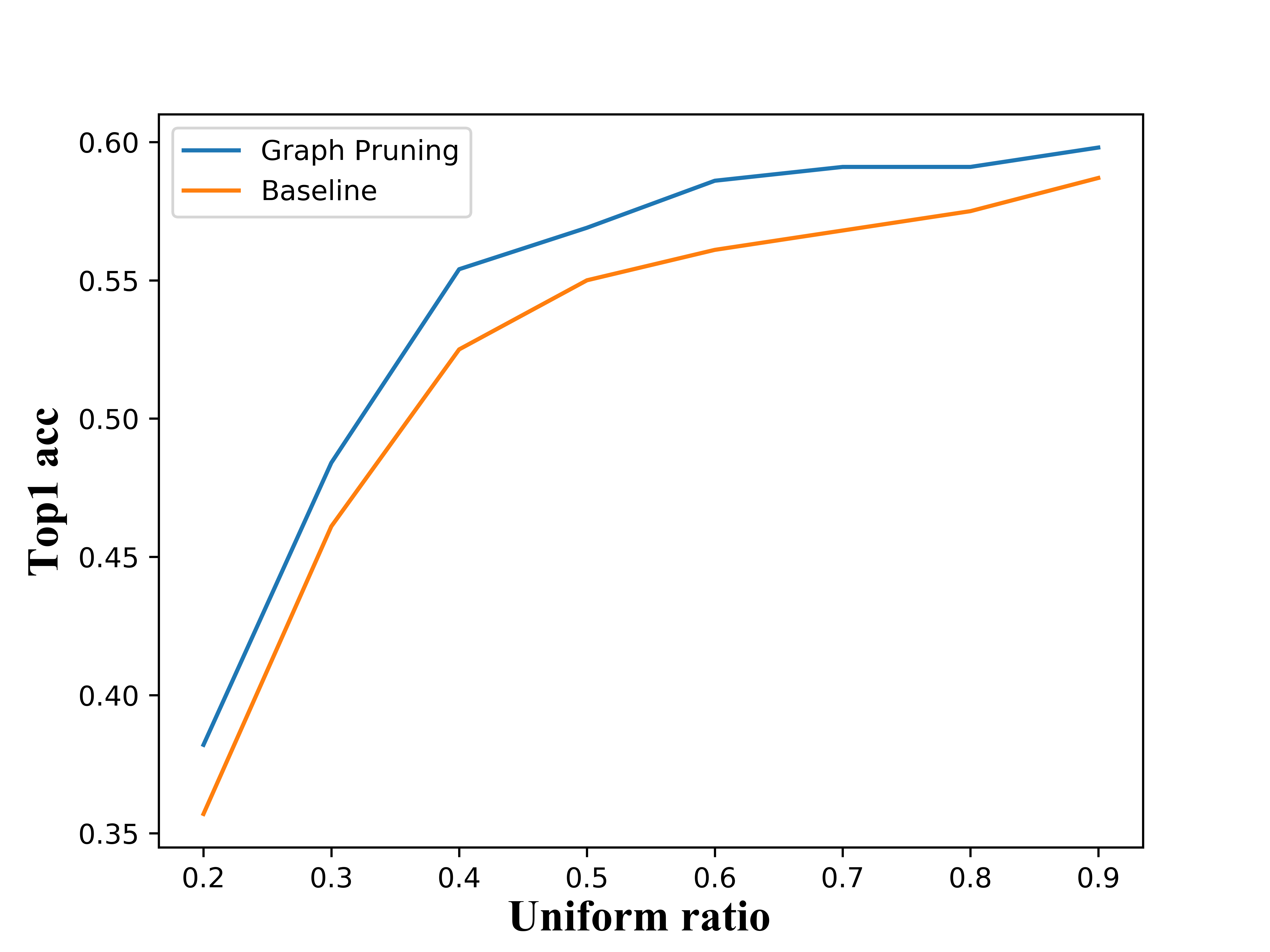}
\end{minipage}}
\subfigure[]{
\begin{minipage}[t]{0.5\linewidth}
\centering
\includegraphics[width=2.4in]{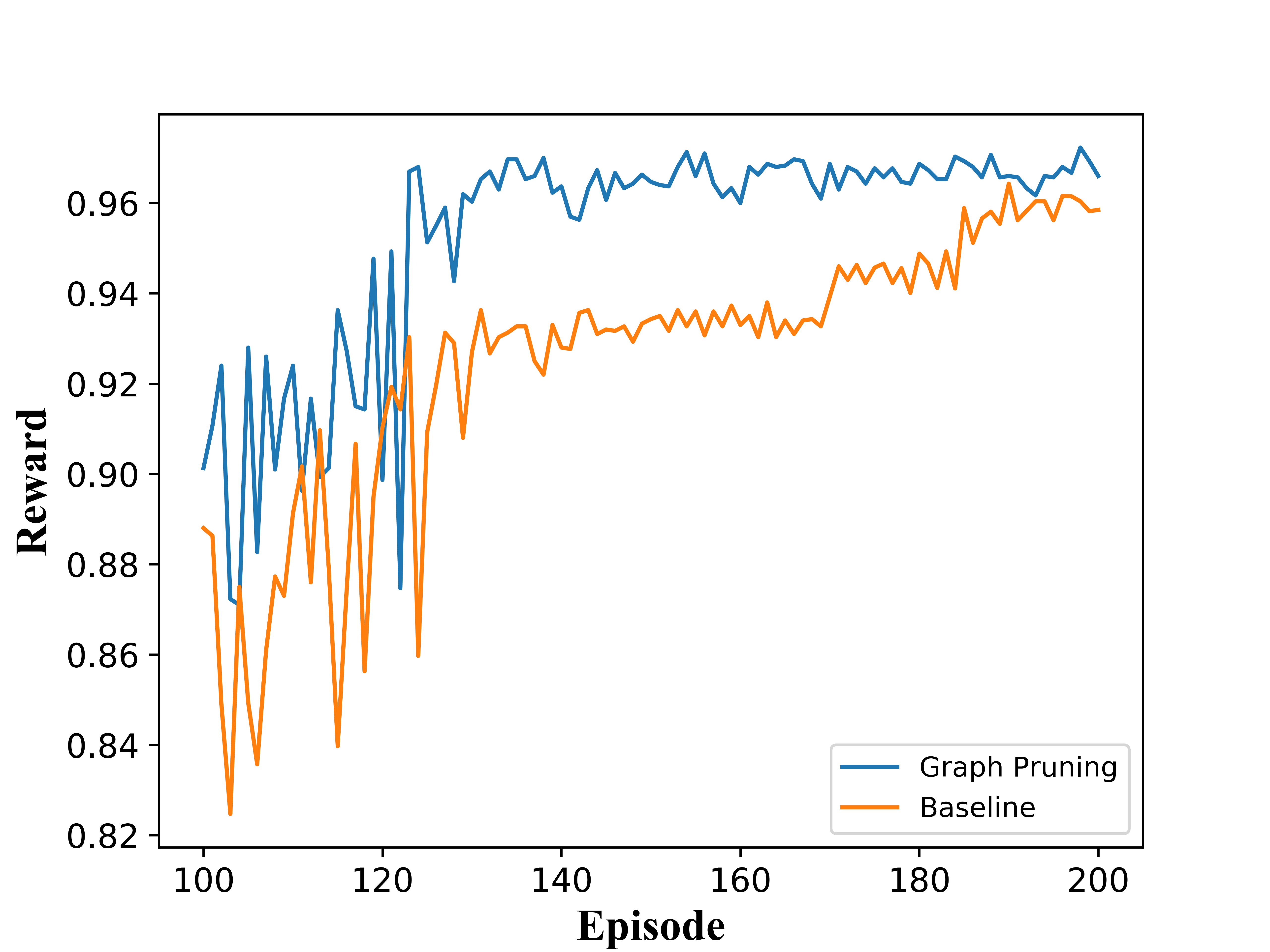}
\end{minipage}}
   \caption{(a) The curve of top-1 accuracy and uniform ratio. (b) The curve of reward and episode.}
\label{fig:ablation}
\end{figure}

In this part, we demonstrate the effectiveness of GraphPruning by comparison on two stages: 
\begin{itemize}
\item \textbf{Training stage} PruningNet \cite{b9} is utilized as the baseline. We compare the performance of PruningNet with and without graph aggregator by inferring the accuracy of several uniformly pruned networks of MobileNet V2 \cite{b2}. The result is shown in \textbf{Fig.\ref{fig:ablation}(a)}. It is found that PruningNet with graph aggregator can achieve much higher accuracy than that without graph aggregator.
\item \textbf{Searching stage} AMC \cite{b8} is utilized as the baseline which does not train PruningNet in advance. A graph aggregator is applied to provide a state for the DDPG agent. The experiment of this part is conducted on the CIFAR-10 dataset. All experimental setting is followed as \cite{b8}. As \textbf{Fig.\ref{fig:ablation}(b)} shows, within graph aggregator, the DDPG agent can get convergence faster.
\end{itemize}
\section{Conclusion}
In this paper, the correlation is analyzed between different layers and finds that some filters in different layers also have a strong correlation. To consider neighboring information in pruning, GraphPruning that uses GCN as the graph aggregator is introduced. Experiments demonstrate that the proposed graph aggregator can effectively extract information for each layer from its neighborhood. Compared with other AutoML methods on ImageNet-2012, the proposed method achieves better or comparable results. In the feature work, GraphPruning can be applied into many CNN-based realtime tasks, i.e., traffic flow management\cite{b46}, pedestrian attribute recognition\cite{b48} and car detection\cite{b49}.

\section*{Acknowledgement}
\label{sub:acknowledgement}
We gratefully acknowledge the support of National Key R\&D Program of China (2018YFB1308400) and Natural Science Foundation of Zhejiang Province (NO. LY21F030018).

{\small
\bibliographystyle{ieee_fullname}
\bibliography{egbib}
}

\end{document}